\documentclass[final]{cvpr}

\usepackage{times}
\usepackage{epsfig}
\usepackage{graphicx}
\usepackage{amsmath}
\usepackage{amssymb}

\usepackage{tabularx}
\usepackage{soul}
\usepackage{xcolor}
\usepackage{multicol}

\usepackage{wrapfig}

\usepackage{caption}
\usepackage{multirow}
\usepackage{amssymb}

\usepackage{tabularx}
\usepackage{soul}
\usepackage{bm}
\usepackage{wrapfig}
\usepackage{color}
\usepackage{amsmath}
\usepackage{multirow}

\usepackage{here}

\usepackage{multicol}
\usepackage{mathtools}
\usepackage[utf8]{inputenc} 
\usepackage[T1]{fontenc}    
\usepackage{url}            
\usepackage{booktabs}       
\usepackage{amsfonts}       
\usepackage{nicefrac}       
\usepackage{microtype}      

\usepackage{hyperref}
\hypersetup{pagebackref=true,breaklinks=true,colorlinks,bookmarks=false}

\usepackage{comment}

\def\cvprPaperID{****} 
\def\confYear{CVPR 2021}

\begin{document}

\title{Perceptual Deep Neural Networks: Adversarial Robustness through Input Recreation}

\author{Danilo Vasconcellos Vargas\\
Kyushu University, Japan\\
{\tt\small vargas@inf.kyushu-u.ac.jp}
\and
Bingli Liao\\
Kyushu University, Japan\\
{\tt\small }
\and
Takahiro Kanzaki\\
Kyushu University, Japan\\
{\tt\small }
}

\maketitle


\begin{abstract}
Adversarial examples have shown that albeit highly accurate, models learned by machines, differently from humans, have many weaknesses. However, humans' perception is also fundamentally different from machines, because we do not see the signals which arrive at the retina but a rather complex recreation of them. In this paper, we explore how machines could recreate the input as well as investigate the benefits of such an augmented perception. In this regard, we propose Perceptual Deep Neural Networks ($\varphi$DNN) which also recreate their own input before further processing. The concept is formalized mathematically and two variations of it are developed (one based on inpainting the whole image and the other based on a noisy resized super resolution recreation). Experiments reveal that $\varphi$DNNs and their adversarial training variations can increase the robustness substantially, surpassing both state-of-the-art defenses and pre-processing types of defenses in 100\% of the tests. 
$\varphi$DNNs are shown to scale well to bigger image sizes, keeping a similar high accuracy throughout; while the state-of-the-art worsen up to 35\%.
Moreover, the recreation process intentionally corrupts the input image. Interestingly, we show by ablation tests that corrupting the input is, although counter-intuitive, beneficial.
Thus, $\varphi$DNNs reveal that input recreation has strong benefits for artificial neural networks similar to biological ones, shedding light into the importance of purposely corrupting the input as well as pioneering an area of perception models based on GANs and autoencoders for robust recognition in artificial intelligence.

\end{abstract}

\section{Introduction}

Recent work has revealed that albeit highly accurate, deep neural networks are far from robust \cite{42503}.
The lack of robustness exist even for extremely small perturbations and simple transformations \cite{madry2018towards,Engstrom2017ARA,su2019one}.
A wide range of defenses were proposed in recent years \cite{43405,grosse2017statistical,li2017adversarial,metzen2017detecting,ma2018characterizing}. 
However, most of them have shortcomings such as relying on obfuscated gradients \cite{obfuscated-gradients} or being biased by the type of perturbation used to train (e.g., adversarial training) \cite{kurakin2017adversarial,Kannan2018adversarial}.

Humans are less affected by small changes in the input.
Interestingly, this is true even when part of the input is completely removed; which happens every second.
Each of our eyes have a blind spot\footnote{The blind spot in each eye is where the optic nerve passes through the optic disc and therefore no photoreceptor cells are present.} where light cannot be perceived. 
Albeit this limitation, when we close one eye we do not see a black spot but a completely filled perception of an image \cite{de1995responses,komatsu2006neural}.
This is an example of how the brain is always predicting what it is viewing, revealing that biological perceptual systems are active rather than passive \cite{mcclelland1981interactive}.
Thus, the images we see every second is rather a creation than mere signals that arrived in the brain, also called perception filling-in and related to predictive coding \cite{clark2013whatever, rao1999predictive,ehinger2017humans}.
In this context, we raise the following question:
\begin{align*}
    \textit{Could deep neural networks also benefit }\\
    \textit{from actively creating its own input?}
\end{align*}

\begin{figure*}[t]
 \centering
 \includegraphics[width=0.9\linewidth,bb=0 0 1920 1080]{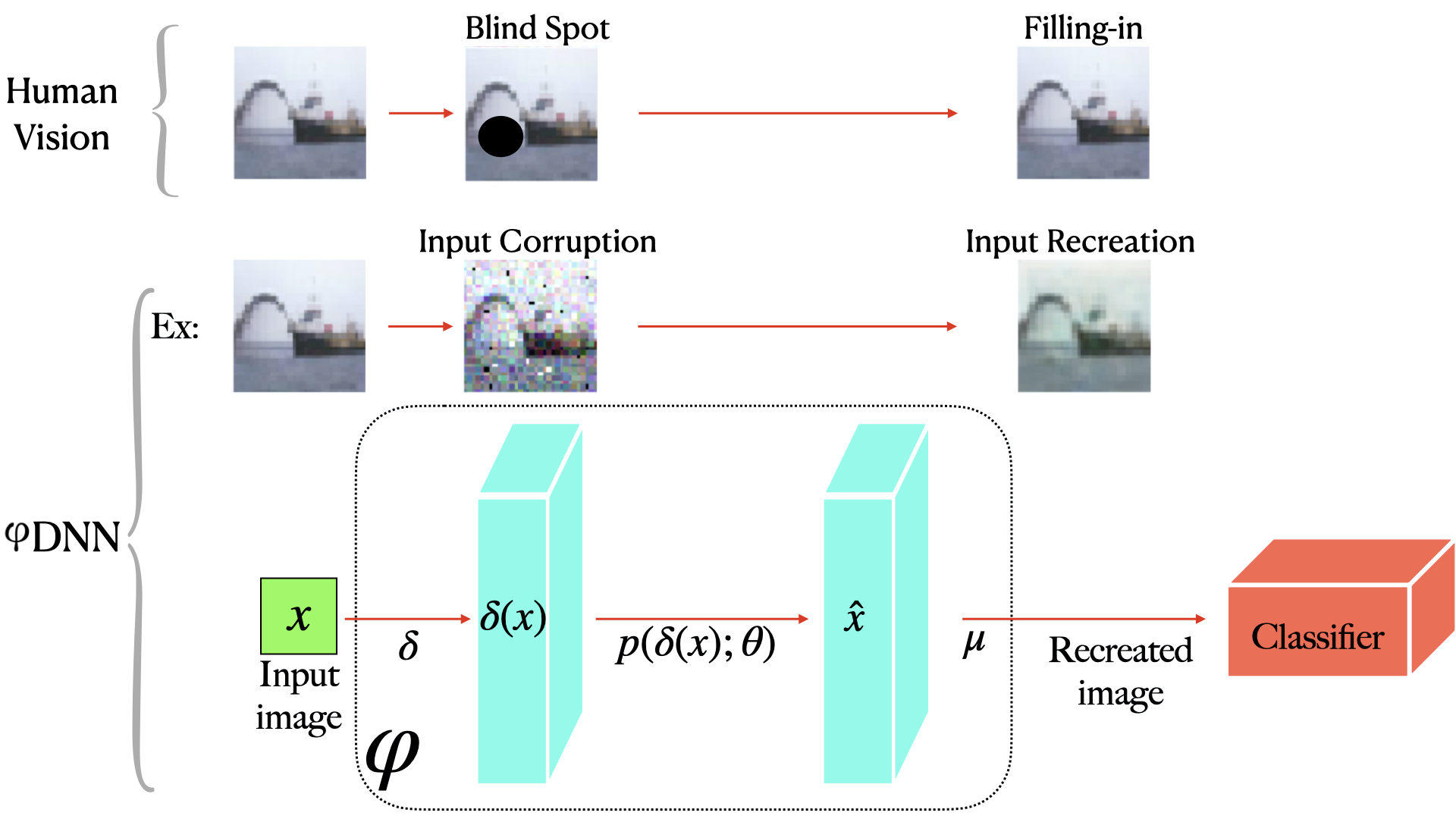}
 \caption{Illustration of the proposed $\varphi$DNN architecture and its similarity to the filling-in phenomena in human vision. Input $x$ is initially corrupted by $\delta(x)$, removing some information while keeping contextual clues.  $p(\delta(x);\theta)$ uses this corrupted image to predict a partial (or whole) recreation 
 which is then aggregated with function $\mu$ to compose a complete recreated image. This recreated image is sent to the vanilla classifier.}
 \label{fig:pred}
\end{figure*}

To answer the question above we developed two perceptual systems that recreate the input image with predictions of it.
One is based on inpainting all parts of the image while the other is based on recreating a super resolution of the image and then resizing it (Section~\ref{sec:phiDNN}).
The recreated input is then fed to a deep neural network which has no access to the original input (Figure~\ref{fig:pred}).
Attacks on both systems suggest that by recreating the input, robustness against adversarial attacks increase.
Furthermore, the input recreation is not mutually exclusive with many of the previous defenses.
It can be used together with adversarial training, for example, to improve further robustness.

\subparagraph{Our contributions.}In this paper, we present input recreation as a novel paradigm to enhance robustness against adversarial samples.
The key contributions can be summarized as follows:
\begin{itemize}
    \item We introduce deep neural networks (DNN) that recreate their own input based solely on contextual hints, called perceptual DNNs ($\varphi$DNN). We describe $\varphi$DNNs formally and conduct experiments on two different implementations of it. 
    \item We propose an inpainting based $\varphi$DNN. It works by predicting removed parts of the image and then joining the predicted parts together into a single completely recreated image. This recreated image is then used as input to a DNN.
    \item We propose a super resolution based $\varphi$DNN which recreates a higher resolution version of the input excluding at the same time any noise present in the original one. The image is later resized and inputted to a DNN.
    \item The results suggest that approaches with active perceptual systems recreating their own input can achieve higher robustness than their counterparts. This is true not only for the best performing system but most of its numerous variations, revealing a strength of the approach. Moreover, $\varphi$DNNs can be used jointly with other defenses to increase robustness further.
    \item Experiments reveal that, for neural networks able to recreate their own input, always purposely corrupting the input (for both training and testing) is mostly beneficial. 
\end{itemize}

\section{Related Works}


\subparagraph{Attack Methods.} 
In this paper, we make use of attack methods for the sole purpose of evaluating the robustness of defenses and neural network models.
Several attack models have been proposed in recent studies. They can be broadly categorized into white box \cite{42503,43405,madry2018towards,carlini2017towards} and black box attacks \cite{papernot2017practical, Brendel2017a,ilyas2018blackbox,tu2019autozoom,dong2019efficient}. 
Many white box models can be summarized as follows. Given a target classifier $C$ and an input pair ($x$, $y$). Let $\mathcal{L}$ be the adversarial loss for the classifier $C$($x'$)  e.g., the cross-entropy loss, and the $\ell_p$ norm used to measure the distance between the legitimate input $x$ and the adversarial input $x'$. Generally, white box attack methods have been proposed by solving the constrained optimization problem:
\begin{equation}
    \underset{x'}{\mathsf{min}} \hspace{0.5em} \mathcal{L}(C(x'), y),\quad \mathsf{s.t.}\quad \parallel x - x'\parallel_p \leq \epsilon.
\end{equation}
Examples of white box attacks are FGSM \cite{43405}, one of the earliest white box attacks, which uses one-step approach to determine the direction to change the pixel value,  and an improved method called projected gradient descent (PGD) with a multiple-step variant  \cite{madry2018towards}.
In contrast, black box attacks have been proposed under more critical and practical conditions with the trade-off of being slower.  
Here, we are also interested in black box attacks which are not based on estimating gradients and therefore can find adversarial samples even when the gradient is masked \cite{obfuscated-gradients}. 
Therefore, tests with more straightforward black box attack methods based on evolutionary strategy such as the one-pixel attack and few-pixel attack fits the purpose \cite{su2019one}. 


\begin{figure*}[t]
 \centering
 \includegraphics[width=0.8\linewidth,bb=0 0 1123 679]{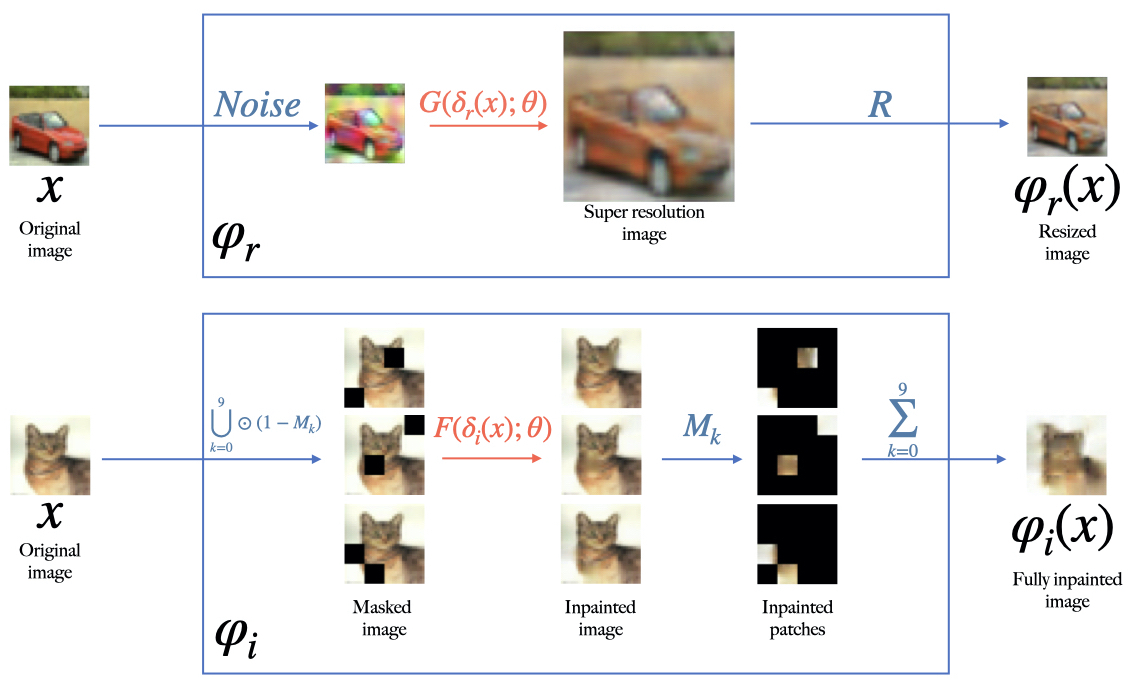}
 \caption{Illustration of the two implementations of $\varphi$DNNs proposed: Noisy Super-resolution Recreation (top) and Full Inpainting Recreation (bottom). 
 }
 \label{fig:cat}
\end{figure*}

\subparagraph{Defenses to Adversarial Attacks.}
Recent studies have proposed various defense mechanisms against the threat of adversarial attacks. Albeit recent efforts, there is not yet a completely effective method. 
Defensive distillation, for example, proposed a smaller neural network which squeezed the content learned by the original one \cite{papernot2016distillation}, however, it was shown to lack robustness in a later paper \cite{carlini2017towards}.
Adversarial training which was firstly proposed by Goodfellow \emph{et al.} \cite{43405} increases the robustness by adding adversarial examples to the training set \cite{huang2015learning}, \cite{madry2018towards}. 
Similarly, adversarial training was also shown vulnerable to attacks in \cite{tramer2018ensemble}.
Other defenses include pre-processing defenses such as the feature squeezing (FS) and spatial smoothing (SS) \cite{xu2017feature}. The objective here is to remove adversarial perturbation in a pre-processing stage.
Recently there are a huge number of defenses proposed, however, they use mostly variations of gradient masking to avoid being attacked which do not confer greater security \cite{obfuscated-gradients}. 
Regarding GAN based defenses, Defense-GAN \cite{samangouei2018defensegan} is based on training a generative adversarial network (GAN) to learn the distribution of original images. Each input would then be used to search for the closest projected input image learned by the generator before proceeding to classification.
One of the main shortcomings is that the distribution learned by the generator is strictly limited by the training data set and the input image might be mapped into an illegitimate space. 
Albeit using GANs in our proposed approach, it shares no other similarities to Defense-GAN. Here, GANs predict parts of the input using the contextual information present, and only after the input has been purposely corrupted.

\subparagraph{Predictive Coding.} Although $\varphi$DNNs do not necessarily use many of the components of predictive coding, it is loosely based on it.
Predictive coding is a theory in neuroscience which postulates that the brain achieves high visual robustness by dynamically updating and predicting neural activities from the environment \cite{pennartz2015brain}. 
Previous studies have shown that the brain uses similar representations to CNNs, but CNNs are not as robust as the brain \cite{cichy2016comparison,wen2018neural}. 
Though there is still no perfect theoretical explanation for how it works, biological plausible models describe it as a recurrently connected hierarchical neural networks \cite{sporns2004small}. Recent research on predictive coding based CNNs imitating the feedforward, feedback, and recurrent connections performed well in object recognition tasks \cite{pmlr-v80-wen18a}.  

This work makes use of both Generative Adversarial Network (GAN) and AutoEncoder (AE) to recreate the images. 
They are described briefly as follows. 
\subsection{Generative Adversarial Network}

Generative Adversarial Network (GAN) is a powerful generative model that consists of two neural networks: a generator network which learns the probability distribution of the input and a discriminator network which distinguishes between generated data and the input  data \cite{goodfellow2014generative}. 

\subparagraph{Super-resolution GAN (SRGAN).}
Super-resolution GAN (SRGAN) generates a photo-realistic high-resolution (HR) image from its downsampled low-resolution (LR) input image. In \cite{ledig2017photo}, they used VGG-19 network to extract high dimension features and designed an alternative function, the perceptual loss function, which consists of content loss and adversarial loss to solve the following min-max optimization problem:
\begin{eqnarray}
    \underset{G}{min}\hspace{0.5em} \underset{D}{max}\hspace{0.5em} \mathbb{E}_{I^{HR}\sim p_{train}(I^{HR})}\lbrack \mathsf{log} D(I^{HR})\rbrack  \nonumber \\
    + \mathbb{E}_{I^{LR}\sim P_G(I^{LR})}\lbrack \mathsf{log}(1-D(G(I^{LR})))\rbrack.
\end{eqnarray}
Here, the generative model \emph{G} maps a given LR input \emph{$I^{LR}$} to its HR counterpart \emph{$I^{HR}$}. The discriminator \emph{D} is trained to distinguish between the produced \emph{$I^{HR}$} images from real inputs.

\subsection{Autoencoder based Inpainting} 
Inpainting is defined as the synthesis of content to fill missing image parts. 
Here, we use an AE to predict the missing pixels with a simple UNET-like architecture \cite{ulyanov2018deep,ronneberger2015u}. Let a masked image $x_0$ be represented as $x_0 = x \odot (1-M)$, in which $M$ is a binary mask, $x$ is the original input image and $\odot $ is the element-wise product operation. Inpainting can be formulated as the following energy minimization problem:
\begin{align*}
 &\underset{\theta}{min}\hspace{0.5em} E(F(x_0;\theta);x), \\
 &E(F(x_0;\theta); x)=|(F(x_0;\theta)-x)|,
\end{align*}
where $F$ is the resulting function from the AE with parameters $\theta$.


\section{$\varphi$DNNs}

In this section, we describe formally the $\varphi$DNN architecture, its motivation as well as two different implementations of it.

\subsection{Technical Motivation Behind Input Recreation}
Beyond the bio-inspired aspect, there are some technical importance for recreating the input in a similar way to humans and other animals.
First, by recreating the input, the neural network and not the environment defines which input will be responsible for the output of the system.
This type of actively modified input provides further control of the input to avoid contextual problems or other issues beyond adversarial samples.
Second, it is now possible to constrain the probability distribution of the input further. This can be done in many ways and is only slightly explored here with added noise. 
Third, with perceptual changes happening all the time, attacking becomes a time-varying function which might be impossible to repeat. This would make calculated attacks near impossible.
Fourth, when facing $\varphi$DNNs, the attacker has less information about the network for he/she does not know even the input now. 
Lastly, gradient-based and gradient estimation based approached tend to perform poorly if the input is changed substantially by the recreation process.


\subsection{$\varphi$DNN's Architecture}
\label{sec:phiDNN}
Consider the perceptual tuple $\varphi$ and its respective function $\varphi(x)$ as follows:
\begin{eqnarray}
    \varphi \coloneqq <\delta, p(\delta(x);\theta), \mu>,\\
    \varphi(x) = \mu(p(\delta(x);\theta)),
\end{eqnarray}
where $\delta$ is a function that corrupts the input, removing some information from it and returning one or multiple corrupted images; $p(\delta(x);\theta)$ is the probability distribution learned by a model that predicts $x$ from the corrupted input $\delta(x)$ based on its learned weights $\theta$; and $\mu$ is the aggregation function which joins partial recreations (when present) into a single recreated image.

$\varphi$DNN is defined as follows:
\begin{equation}
\varphi\textbf{DNN} \coloneqq C(\varphi(x)),
\end{equation}
in which $C$ is a classifier that receives as input the output from the perceptual function $\varphi(x)$. 

\begin{figure}[h]
 \centering
 \includegraphics[width=7cm,bb=0 0 930 265]{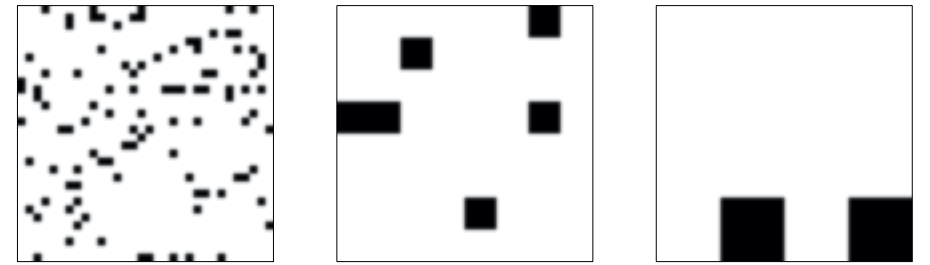}
 \caption{Three masks created for FIR with grid size of respectively (from left to right) $1$, $4$ and $8$.}
 \label{fig:3mask}
\end{figure}

\begin{table*}[ht]
 \begin{center}
  \caption{Comparison between Feature Scattering (the current state-of-the-art defense) and the two proposed $\varphi$DNNs with a simple adversarial training (NSR$_{adv}$ and FIR$_{adv}$) on CIFAR10, SVHN and Imagenette. Results show the accuracy of defenses under attack. For reference, we include the $\varphi$DNNs without adversarial training (NSR and FIR), only ResNet with the same simple adversarial training used on NSR$_{adv}$ and FIR$_{adv}$ (ResNet$_{adv}$); as well as vanilla ResNet. ResNet$_{pruned}$ means the trained ResNet is pruned 50\%, retrained and then ultimately pruned to 80\%.}
  
  \begin{tabularx}{0.9\linewidth}{Xc|cc|cc} \hline
   Defense & test acc & 1px attack & 10px attack &
   \begin{tabular}{c}
    FGSM\\($\epsilon=8$) 
   \end{tabular}
   &\begin{tabular}{c}
   PGD\\($\epsilon=8$)
   \end{tabular}
   \\ \hline     
   \multicolumn{6}{c}{Imagenette} \\
  \hline

   \textbf{Ours:} NSR$_{adv}$& 77.0 & 75.3 & 75.6 & 51.6 & 46.7 \\
   \textbf{Ours:} FIR$_{adv}$& 78.1 & 73.7 & 62.9 & \textbf{74.8} & \textbf{66.2} \\
   FScattering & 72.4 & 59.5 & 52.2 & 42.3 & 43.2 \\
   \hline \hline
   NSR & 81.7 & 77.5 & \textbf{77.6} & 37.4 & 45.8 \\
   FIR & 85.4 & \textbf{81.1} & 76.9 & 5.0 & 0.3 \\

   \hline
     \multicolumn{6}{c}{CIFAR10} \\
  \hline
  
   \textbf{Ours:} NSR$_{adv}$& 82.7 & \bf{78.1} & \bf{73.9} & 75.3 & {75.5} \\
   \textbf{Ours:} FIR$_{adv}$& 88.5 & 73.6 & 50.4 & \bf{88.3} & \bf{82.5} \\
   FScattering & 90.0 & 72.2 & 47.7 & 78.4 & 70.5 \\
   TRADES & 87.4 & --- & --- & --- & 51.6 \\
   LLR & 86.8 & --- & --- & --- & 54.2 \\
   \hline \hline
   NSR & 83.8 & 78.3 & 73.9 & 63.4 & 69.3 \\
   FIR  & 89.4 & 78.1 & 54.2 & 25.6 & 5.0 \\
   ResNet$_{adv}$ & 89.2 & 60.3 & 17.8 & 25.2 & 0.0 \\
   ResNet  & 93.0 & 62.1 & 18.2 & 17.7 & 0.0 \\
   ResNet$_{pruned}$& 85.8 & 41.9 & 4.1 & 63.8 & 56.8 \\
   
   \hline   
   \multicolumn{6}{c}{SVHN} \\
  \hline
  
   \textbf{Ours:} NSR$_{adv}$& 92.4 & 89.1 & 87.7 & 85.0 & 86.9 \\
   \textbf{Ours:} FIR$_{adv}$& 93.0 & 80.9 & {53.9} & \bf{93.0} & \bf{91.5} \\
   FScattering & 96.2 & 81.4 & 46.0 & 83.5 & 52.0 \\
   \hline \hline
   NSR & 93.0 & \bf{90.2} & \bf{90.1} & 74.4 & 85.6 \\
   FIR & 96.4 & 80.4 & 57.1 & 49.0 & 7.9\\ 
   
   \hline
  \end{tabularx}
  \label{sota}
 \end{center}
\end{table*}

\subparagraph{Noisy Super-Resolution Recreation (NSR)}
Here we define an implementation of the $\varphi$DNN's architecture using super resolution and images corrupted with noise. 
Note that images are always corrupted with noise (i.e., in both training and testing).
Let $x$ be a given input and $R$ a function which resizes the high resolution image to the original resolution. The process can be defined as follows:
\begin{gather}
 \varphi_r\coloneqq<\delta_r, p_r(\delta(x);\theta_r), \mu_r>, \\
 \quad \mathsf{where}\quad\delta_r = Noise, p_r = G(\delta(x);\theta), \mu_r = R.\nonumber
\end{gather}
note that $Noise$ is an arbitrary noise function which returns a noisy image.
$G(\delta(x);\theta)$ is the generator of SRGAN which maps an image from low resolution to high resolution and tries to clean the always present noise (illustrated in Figure~\ref{fig:cat}).

\subparagraph{Full Inpainting Recreation (FIR)}

To demonstrate that $\varphi$DNNs can be developed in many forms, here, we propose a $\varphi$DNN based on inpainting the whole image.
Specifically, $\varphi_i$ is defined as follows.
 \begin{gather}
  \varphi_i\coloneqq<\delta_i,p_i(\delta(x);\theta_i),\mu_i>, \\
  \quad \mathsf{where} \quad  \delta_i(x)=\bigcup_{k=0}^9\odot(1- M_k) \cdot x, \nonumber \\
  p_i=F(\delta(x);\theta),\mu_i(x)=\sum_{k=0}^9  M_k \cdot x, \nonumber
 \end{gather}
where $M_k$ are masks such that their sum is equal to the identity matrix ($\sum_{k=0}^9  M_k = I$) and their multiplication is equal to 0 ($\prod_{k=0}^9  M_k = 0$). Therefore, each of the masks hide a specific part of the image; and together they mask the whole image. 
$\delta_i(x)$  (i.e., $\bigcup_{k=0}^9\odot(1- M_k)\cdot x$) creates a set with $10$ masked inputs. All masked inputs are then inpainted with $F(\delta(x);\theta)$ and lastly all inpainted parts are joined together through $\sum_{k=0}^9  M_k \cdot x$.
Figure~\ref{fig:cat} shows an illustration of the process.

\begin{table*}[t]
 \begin{center}
  \caption{Comparison of proposed methods with other pre-prossessing based defenses. NSR and FIR models use the best setting from Tables~\ref{nsr_vars} and \ref{fir_vars2} while the other ones use AllConv and the best settings out of a couple of experiments.}
  \begin{tabularx}{0.9\linewidth}{Xc|cc|cc} \hline
   Defense &test acc& 1px attack & 10px attack &
   \begin{tabular}{c}
    FGSM\\($\epsilon=8$) 
   \end{tabular}
   &\begin{tabular}{c}
   PGD\\($\epsilon=8$)
   \end{tabular}
   \\ \hline 
   
   {\bf Ours:} NSR & 83.8 & \bf{78.3 }& \bf{73.9} & \bf{63.4 }& \bf{69.3} \\
   {\bf Ours:} FIR & 89.4 & 73.2 & 54.2 & 25.6 & 5.0 \\ 
   FS & 79.2 & 42.7 & 10.5 & 17.3 & 0.0 \\
   SS & 78.6 & 69.6 & 38.8 & 20.1 & 0.9 \\
   JPEG & 73.0 & 21.0 & 1.5 & 31.5 & 5.4 \\
   LS & 91.4 & 48.6 & 15.0 & 50.5 & 0.0 \\
   
   \hline
  \end{tabularx}
  \label{prepro}
 \end{center}
\end{table*}

\section{Experiments}

To evaluate $\varphi$DNN architecture, we test here the robustness of two implementations of it (i.e., FIR and NSR) by attacking them with different types of attacks.
The proposed architecture is also compared with other defenses in three datasets (CIFAR, SVHN and a subset of Imagenet called Imagenette \cite{shleifer2019using}).

To evaluate the robustness of systems avoiding biases and the sole presence of gradient masking, we employ two white box attacks (FGSM and PGD) as well as non-gradient based black box attacks (one pixel and ten pixel attack).
In this paper, every attack is repeated for $500$ uniformly sampled random images of the test data set with the average attack accuracy being reported. 
For all experiments, the CIFAR-10 dataset is normalized to the range $[0, 1]$.
The machines used in the experiments are equipped with NVIDIA GeForce RTX 2080 Ti and AMD Ryzen 9 3950x 16-core.

Regarding FIR, $10$ masks are created, each of them removing $10\%$ of the image. (Figure~\ref{fig:3mask}).
To create the masks, a grid of a given size is set over the 32$\times$32 image and then multiple pieces of this grid are randomly selected to form one mask. Pieces are selected until $10\%$ of the image is covered.
The inpainting model is trained with a corresponding mask size covering $10\%$ of the image and with epochs and batch size of respectively $20$ and $32$.

Regarding NSR, to create the training dataset for SRGAN we resize CIFAR-10 dataset to 128$\times$128 as the high resolution ground truth and add noise to the training dataset during training.
The type of noise used is bi-linear interpolation for both up and downsizing.
In order to match with the normalization, we replace tanh with sigmoid as the activation function for the last convolution layer in SRGAN's generator. We train SRGAN with $1000$ epochs and set batch size to $20$ to ensure convergence.


\subsection{Comparison with other Defenses}

Table~\ref{sota} compares the last development in adversarial training, i.e. Feature Scaterring \cite{feature_scatter}, with the proposed algorithms and variations of them trained with a simple adversarial training. Other state-of-the-art defenses such as LLR and TRADES are also included in some of tests, using the original results reported from their papers \cite{qin2019adversarial,zhang2019theoretically}.

Results show that both $\varphi$DNNs with adversarial training surpass FScattering for all of the attacks (the only exception is the SVHN test in which FIR$_{adv}$ gets $80.9$\% against $81.4$\% achieved by FScattering). 
It is known that adversarial training methods such as FScattering perform poorly when the attacking distribution differ from the data used to learn. This applies to FScattering as well which can be attacked with more than 50\% attack accuracy with 10px attack.
Having said that, it is impressive that both NSR and FIR can surpass FScattering even on FGSM and PGD which are close to the augmented distribution of noisy images FScattering used to learn.
Notice that the same adversarial training that has little change on the vanilla Resnet (i.e., ResNet$_{adv}$) is very effective on NSR and FIR. For example, in CIFAR under a PGD attack, FIR$_{adv}$ is 82.5\% accurate against a 5\% accuracy of the vanilla FIR and a 0\% accuracy of the same adversarial training applied on a vanilla ResNet.
Thus, it is expected that if a state-of-the-art adversarial training is applied to NSR and FIR, their robustness should improve even further.
Pruned networks were also added to demonstrate that lower accuracy pruned neural networks are not comparable with current defenses.

In fact, if we take into account that FScattering and $\varphi$DNN are (a) different in nature and (b) can be also used together. 
It can be justified that $\varphi$DNNs should be compared with other pre-processing defenses and not adversarial training ones. 
We follow this rational and compare the proposed methodology in Table~\ref{prepro} with other pre-processing defenses such as FS, SS, JPEG compression defence (JPEG) \cite{das2017keeping} and Label Smoothing (LS) \cite{hazan2016perturbations}.
All defenses used ResNet with the same type of augmentation.
Note that we also tried to include DefenseGAN but it failed to learn properly on CIFAR10.

Both $\varphi$DNNs surpass all others in most of the attacks. 
The result is expected since $\varphi$DNNs do not only pre-process images, they recreate them based on contextual information and previous learned distribution.
The only exception is FIR for FGSM and PGD (Table~\ref{prepro}).
FIR's poor results on PGD and FGSM are less obvious. It is related to the grid size which is discussed in Section~\ref{fir_analysis}.

\begin{table*}[htb]
 \begin{center}
  \caption{Attack accuracy for both NSR and SR (NSR without the added noise $\delta_r()$) trained with different types of noise and connected to ResNet. We tested Gaussian noise with 0 mean ($\mu$), and variances ($\sigma^{2}$) of 0.01. For Panda noise, the scalar number (0.01) represents the probability ($\alpha$ and $\beta$) of white and black pixels present in the image. $A+B$ represents that two types of noises $A$ and $B$ are summed together. The subscript $_T$ means that the classifier was retrained with a data set made of recreated images (i.e., images from $\varphi_r(x)$).}
  \begin{tabularx}{0.8\linewidth}{ccc|c|cc} \hline
    Defense&Noise& test acc & 10px attack &
   \begin{tabular}{c}
    FGSM\\($\epsilon=8$) 
   \end{tabular}
   &\begin{tabular}{c}
   PGD\\($\epsilon=8$)
   \end{tabular}
   \\ \hline
   
   NSR &+ResNet[Guassian0.01]& 79.2 & \bf{60.5} & \bf{64.5} & 68.1 \\
   SR &+ResNet[Guassian0.01]& 77.4 & 16.2 & 61.5 & \bf{68.9} \\
   \hline \hline
   NSR &+ResNet[Panda0.01]& 91.0 & \bf{49.1} & \bf{70.8} & \bf{67}.2 \\
   SR &+ResNet[Panda0.01]& 91.5 & 30.8 & 25.3 & 11.0 \\
   \hline \hline
   NSR &+ResNet[Guassian+Panda]& 77.0 & \bf{64.5} & \bf{62.8} & 66.7 \\
   SR &+ResNet[Guassian+Panda]& 81.9 & 37.9 & 61.4 & \bf{70.6} \\
   \hline \hline 
   NSR &+ResNet[Guassian+Panda]$_{T}$& 83.8 & \bf{74.6} & 63.4 & 69.3 \\
   SR &+ResNet[Guassian+Panda]$_{T}$& 84.5 & 44.6 & \bf{65.9} & \bf{73.0}
   
   \\ \hline
  \end{tabularx}
  \label{nsr_vars}
 \end{center}
\end{table*}

\begin{table*}[htb]
 \begin{center}
  \caption{Comparing the difference of grid size on FIR's accuracy and robustness. ResNet is the vanilla classifier while FIR$^+_1$, FIR$^+_4$ and FIR$^+_8$ means using ResNet in the FIR's architecture with grid size of respectively $1$, $4$ and $8$. Each inpainting model is trained with the corresponding grid size only, and the classifier model is trained with corresponding inpainting image from $\varphi_i(x)$. }
  \begin{tabularx}{0.8\linewidth}{Xcccccc} \hline
    & Test accuracy & 1px attack & 10px attack&
    \begin{tabular}{c}
    FGSM\\($\epsilon=8$) 
   \end{tabular}
   &\begin{tabular}{c}
   PGD\\($\epsilon=8$)
   \end{tabular}
    \\ \hline 
    
   ResNet & 93.0 & 56.0 & 18.2 & 17.7 & 0.0 \\\hline
   FIR$_1$ & 89.4 & \bf{73.2} & \bf{54.2} & 25.6 & 5.0 \\
   FIR$_4$ & 82.7 & 69.2 & 43.2 & 48.0 & 49.1 \\
   FIR$_8$ & 73.5 & 59.0 & 25.9 & \bf{49.5} & \bf{53.7}
   
   \\ \hline
  \end{tabularx}
  \label{fir_vars2}
 \end{center}
\end{table*}

\begin{figure*}[h!]
    \centering
    \includegraphics[width=0.7\linewidth]{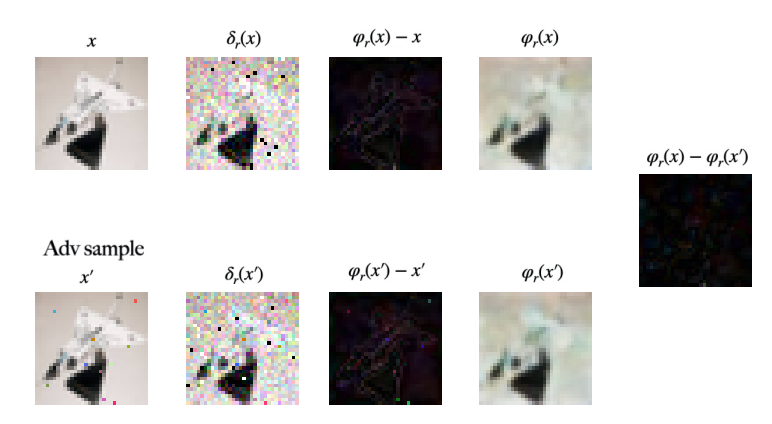}
    \caption{
    Behavior of an image sample $x$ and its respective adversarial sample $x'$ throughout $\varphi_r(x)$. 
    }
    \label{nsr_adv}
\end{figure*}

\begin{figure*}[h!]
 \centering
 \includegraphics[width=\linewidth]{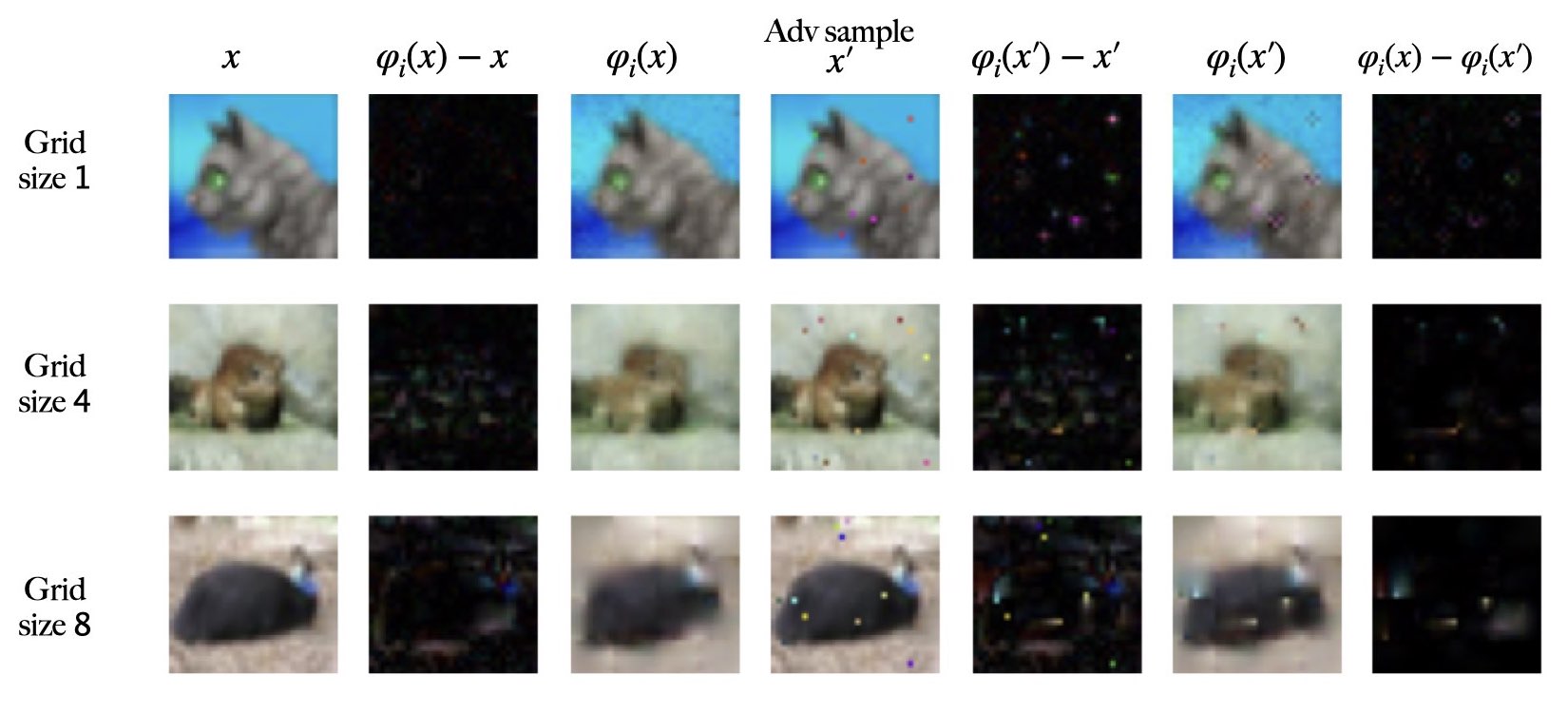}
 \caption{
 Behavior of $\varphi_i(x)$ on an image sample $x$ and its respective adversarial sample $x'$.
 }
 \label{fir_adv}
\end{figure*}
\subsection{NSR Analysis: When Losing Information is Beneficial}

NSR corrupts the input image possibly losing some information.
Here we will investigate if this loss of information has any deleterious consequences.
We will also analyze the behavior of NSR on adversarial samples.
To analyze the influence of the initial input corruption by $\delta_r(x)$, an ablation test is made, in which $\delta_r(x)$ is removed from $\varphi_r(x)$ (this algorithm is called SR in Table~\ref{nsr_vars}).
Results show an increased robustness and similar accuracy.
Specifically, in 8 out of 12 tests, the robustness of NSR surpassed the ablated algorithm SR. 
Regarding the accuracy, both NSR and SR performed similarly.

These results reveal, perhaps counter-intuitively, that always adding noise ($\delta_r(x)$) to the input is mostly beneficial for neural networks that recreate their input.
On average, it usually improves robustness while leaving accuracy unchanged. 
There are two reasons for such a behavior: (a) always adding noise constrains the image distribution to non-smooth pixel transitions and (b) an always changing input is harder to attack.



\subsection{FIR Analysis}
\label{fir_analysis}

In this section, FIR will be analyzed with relation to its grid size.
For L$0$ attacks (1px, 5px and 10px attacks), FIR performs better with lower grid values while higher grid values are better suited to L$\infty$ attacks (PGD and FGSM) (Table~\ref{fir_vars2}).
This is expected since L$0$ attacks perturbs fewer pixels and therefore punctual corrections are better. The opposite is true for L$\infty$.
Attacking FIR is difficult because for a pixel to be modified in the final image $\varphi_i(x)$, many pixels around it must be changed (for lower grid values) or pixels near the mask needs to be changed (for higher grid values). This creates a bigger burden on the attacker and causes many attacks to fail.
A simple version of adversarial training (with FGSM created adversarial samples) improves substantially the advantages of FIR, allowing it to even surpass the state-of-the-art (Table~\ref{sota}).

\section{Conclusions}
In this paper, we proposed $\varphi$DNNs which make use of corrupting functions and context-based prediction to recreate the input.
We showed that, perhaps surprisingly, corrupting the input is beneficial to robustness. 
Moreover, two implementations of $\varphi$DNNs surpassed the state-of-the-art while possessing much better scaling for datasets with bigger image sizes. Note that the implementations used here do not utilize state-of-the-art GANs or AEs as well as only a classic adversarial training scheme. Therefore, state-of-the-art GANs/AEs/adversarial training should increase further the results reported here, which are already surpassing the state-of-the-art in all of the tests.

Thus, this paper proposes a novel paradigm for robust neural networks with state-of-the-art results which should, hopefully, incentive further investigation into $\varphi$DNNs and other perception models. It also opens new paths for robust artificial intelligence, towards safer applications.

\bibliographystyle{plain}
\bibliography{ref}

\begin{thebibliography}{10}

\bibitem{obfuscated-gradients}
Anish Athalye, Nicholas Carlini, and David Wagner.
\newblock Obfuscated gradients give a false sense of security: Circumventing
  defenses to adversarial examples.
\newblock In {\em Proceedings of the 35th International Conference on Machine
  Learning, {ICML} 2018}, July 2018.

\bibitem{Brendel2017a}
W.~Brendel, J.~Rauber, and M.~Bethge.
\newblock Decision-based adversarial attacks: Reliable attacks against
  black-box machine learning models.
\newblock In {\em International Conference on Learning Representations}, 2018.

\bibitem{carlini2017towards}
Nicholas Carlini and David Wagner.
\newblock Towards evaluating the robustness of neural networks.
\newblock In {\em 2017 ieee symposium on security and privacy (sp)}, pages
  39--57. IEEE, 2017.

\bibitem{cichy2016comparison}
Radoslaw~Martin Cichy, Aditya Khosla, Dimitrios Pantazis, Antonio Torralba, and
  Aude Oliva.
\newblock Comparison of deep neural networks to spatio-temporal cortical
  dynamics of human visual object recognition reveals hierarchical
  correspondence.
\newblock {\em Scientific reports}, 6:27755, 2016.

\bibitem{clark2013whatever}
Andy Clark.
\newblock Whatever next? predictive brains, situated agents, and the future of
  cognitive science.
\newblock {\em Behavioral and brain sciences}, 36(3):181--204, 2013.

\bibitem{das2017keeping}
Nilaksh Das, Madhuri Shanbhogue, Shang-Tse Chen, Fred Hohman, Li~Chen,
  Michael~E Kounavis, and Duen~Horng Chau.
\newblock Keeping the bad guys out: Protecting and vaccinating deep learning
  with jpeg compression.
\newblock {\em arXiv preprint arXiv:1705.02900}, 2017.

\bibitem{de1995responses}
Peter De~Weerd, Ricardo Gattass, Robert Desimone, and Leslie~G Ungerleider.
\newblock Responses of cells in monkey visual cortex during perceptual
  filling-in of an artificial scotoma.
\newblock {\em Nature}, 377(6551):731--734, 1995.

\bibitem{dong2019efficient}
Yinpeng Dong, Hang Su, Baoyuan Wu, Zhifeng Li, Wei Liu, Tong Zhang, and Jun
  Zhu.
\newblock Efficient decision-based black-box adversarial attacks on face
  recognition.
\newblock In {\em Proceedings of the IEEE Conference on Computer Vision and
  Pattern Recognition}, pages 7714--7722, 2019.

\bibitem{ehinger2017humans}
Benedikt~V Ehinger, Katja H{\"a}usser, Jose~P Ossandon, and Peter K{\"o}nig.
\newblock Humans treat unreliable filled-in percepts as more real than
  veridical ones.
\newblock {\em Elife}, 6:e21761, 2017.

\bibitem{Engstrom2017ARA}
Logan Engstrom, Dimitris Tsipras, Ludwig Schmidt, and Aleksander Madry.
\newblock A rotation and a translation suffice: Fooling cnns with simple
  transformations.
\newblock {\em ArXiv}, abs/1712.02779, 2017.

\bibitem{goodfellow2014generative}
Ian Goodfellow, Jean Pouget-Abadie, Mehdi Mirza, Bing Xu, David Warde-Farley,
  Sherjil Ozair, Aaron Courville, and Yoshua Bengio.
\newblock Generative adversarial nets.
\newblock In {\em Advances in neural information processing systems}, pages
  2672--2680, 2014.

\bibitem{43405}
Ian Goodfellow, Jonathon Shlens, and Christian Szegedy.
\newblock Explaining and harnessing adversarial examples.
\newblock In {\em International Conference on Learning Representations}, 2015.

\bibitem{grosse2017statistical}
Kathrin Grosse, Praveen Manoharan, Nicolas Papernot, Michael Backes, and
  Patrick McDaniel.
\newblock On the (statistical) detection of adversarial examples.
\newblock {\em arXiv preprint arXiv:1702.06280}, 2017.

\bibitem{hazan2016perturbations}
Tamir Hazan, George Papandreou, and Daniel Tarlow.
\newblock {\em Perturbations, Optimization, and Statistics}.
\newblock MIT Press, 2016.

\bibitem{huang2015learning}
Ruitong Huang, Bing Xu, Dale Schuurmans, and Csaba Szepesv{\'a}ri.
\newblock Learning with a strong adversary.
\newblock {\em arXiv preprint arXiv:1511.03034}, 2015.

\bibitem{ilyas2018blackbox}
Andrew Ilyas, Logan Engstrom, Anish Athalye, and Jessy Lin.
\newblock Black-box adversarial attacks with limited queries and information.
\newblock In {\em Proceedings of the 35th International Conference on Machine
  Learning, {ICML} 2018}, July 2018.

\bibitem{Kannan2018adversarial}
Harini Kannan, Alexey Kurakin, and Ian~J. Goodfellow.
\newblock Adversarial logit pairing.
\newblock {\em ArXiv}, abs/1803.06373, 2018.

\bibitem{komatsu2006neural}
Hidehiko Komatsu.
\newblock The neural mechanisms of perceptual filling-in.
\newblock {\em Nature reviews neuroscience}, 7(3):220--231, 2006.

\bibitem{kurakin2017adversarial}
Alexey Kurakin, J.~Ian Goodfellow, and Samy Bengio.
\newblock Adversarial machine learning at scale.
\newblock {\em international conference on learning representations}, 2017.

\bibitem{ledig2017photo}
Christian Ledig, Lucas Theis, Ferenc Husz{\'a}r, Jose Caballero, Andrew
  Cunningham, Alejandro Acosta, Andrew Aitken, Alykhan Tejani, Johannes Totz,
  Zehan Wang, et~al.
\newblock Photo-realistic single image super-resolution using a generative
  adversarial network.
\newblock In {\em Proceedings of the IEEE conference on computer vision and
  pattern recognition}, pages 4681--4690, 2017.

\bibitem{li2017adversarial}
Xin Li and Fuxin Li.
\newblock Adversarial examples detection in deep networks with convolutional
  filter statistics.
\newblock In {\em Proceedings of the IEEE International Conference on Computer
  Vision}, pages 5764--5772, 2017.

\bibitem{ma2018characterizing}
Xingjun Ma, Bo~Li, Yisen Wang, Sarah~M Erfani, Sudanthi Wijewickrema, Grant
  Schoenebeck, Dawn Song, Michael~E Houle, and James Bailey.
\newblock Characterizing adversarial subspaces using local intrinsic
  dimensionality.
\newblock In {\em 6th International Conference on Learning Representations,
  ICLR 2018}, 2018.

\bibitem{madry2018towards}
Aleksander Madry, Aleksandar Makelov, Ludwig Schmidt, Dimitris Tsipras, and
  Adrian Vladu.
\newblock Towards deep learning models resistant to adversarial attacks.
\newblock In {\em International Conference on Learning Representations}, 2018.

\bibitem{mcclelland1981interactive}
James~L McClelland and David~E Rumelhart.
\newblock An interactive activation model of context effects in letter
  perception: I. an account of basic findings.
\newblock {\em Psychological review}, 88(5):375, 1981.

\bibitem{metzen2017detecting}
Jan~Hendrik Metzen, Tim Genewein, Volker Fischer, and Bastian Bischoff.
\newblock On detecting adversarial perturbations.
\newblock In {\em Proceedings of 5th International Conference on Learning
  Representations (ICLR)}, 2017.

\bibitem{papernot2017practical}
Nicolas Papernot, Patrick McDaniel, Ian Goodfellow, Somesh Jha, Z~Berkay Celik,
  and Ananthram Swami.
\newblock Practical black-box attacks against machine learning.
\newblock In {\em Proceedings of the 2017 ACM on Asia conference on computer
  and communications security}, pages 506--519, 2017.

\bibitem{papernot2016distillation}
Nicolas Papernot, Patrick McDaniel, Xi~Wu, Somesh Jha, and Ananthram Swami.
\newblock Distillation as a defense to adversarial perturbations against deep
  neural networks.
\newblock In {\em 2016 IEEE Symposium on Security and Privacy (SP)}, pages
  582--597. IEEE, 2016.

\bibitem{pennartz2015brain}
Cyriel~MA Pennartz.
\newblock {\em The brain's representational power: on consciousness and the
  integration of modalities}.
\newblock MIT Press, 2015.

\bibitem{qin2019adversarial}
Chongli Qin, James Martens, Sven Gowal, Dilip Krishnan, Krishnamurthy
  Dvijotham, Alhussein Fawzi, Soham De, Robert Stanforth, and Pushmeet Kohli.
\newblock Adversarial robustness through local linearization.
\newblock In {\em Advances in Neural Information Processing Systems}, pages
  13847--13856, 2019.

\bibitem{rao1999predictive}
Rajesh~PN Rao and Dana~H Ballard.
\newblock Predictive coding in the visual cortex: a functional interpretation
  of some extra-classical receptive-field effects.
\newblock {\em Nature neuroscience}, 2(1):79--87, 1999.

\bibitem{ronneberger2015u}
Olaf Ronneberger, Philipp Fischer, and Thomas Brox.
\newblock U-net: Convolutional networks for biomedical image segmentation.
\newblock In {\em International Conference on Medical image computing and
  computer-assisted intervention}, pages 234--241. Springer, 2015.

\bibitem{samangouei2018defensegan}
Pouya Samangouei, Maya Kabkab, and Rama Chellappa.
\newblock Defense-{GAN}: Protecting classifiers against adversarial attacks
  using generative models.
\newblock In {\em International Conference on Learning Representations}, 2018.

\bibitem{shleifer2019using}
Sam Shleifer and Eric Prokop.
\newblock Using small proxy datasets to accelerate hyperparameter search.
\newblock {\em arXiv preprint arXiv:1906.04887}, 2019.

\bibitem{sporns2004small}
Olaf Sporns and Jonathan~D Zwi.
\newblock The small world of the cerebral cortex.
\newblock {\em Neuroinformatics}, 2(2):145--162, 2004.

\bibitem{su2019one}
Jiawei Su, Danilo~Vasconcellos Vargas, and Kouichi Sakurai.
\newblock One pixel attack for fooling deep neural networks.
\newblock {\em IEEE Transactions on Evolutionary Computation}, 23(5):828--841,
  2019.

\bibitem{42503}
Christian Szegedy, Wojciech Zaremba, Ilya Sutskever, Joan Bruna, Dumitru Erhan,
  Ian Goodfellow, and Rob Fergus.
\newblock Intriguing properties of neural networks.
\newblock In {\em International Conference on Learning Representations}, 2014.

\bibitem{tramer2018ensemble}
Florian Tram{\`e}r, Alexey Kurakin, Nicolas Papernot, Ian Goodfellow, Dan
  Boneh, and Patrick~Drew McDaniel.
\newblock Ensemble adversarial training: Attacks and defenses.
\newblock In {\em 6th International Conference on Learning Representations,
  ICLR 2018}, 2018.

\bibitem{tu2019autozoom}
Chun-Chen Tu, Paishun Ting, Pin-Yu Chen, Sijia Liu, Huan Zhang, Jinfeng Yi,
  Cho-Jui Hsieh, and Shin-Ming Cheng.
\newblock Autozoom: Autoencoder-based zeroth order optimization method for
  attacking black-box neural networks.
\newblock In {\em Proceedings of the AAAI Conference on Artificial
  Intelligence}, volume~33, pages 742--749, 2019.

\bibitem{ulyanov2018deep}
Dmitry Ulyanov, Andrea Vedaldi, and Victor Lempitsky.
\newblock Deep image prior.
\newblock In {\em Proceedings of the IEEE Conference on Computer Vision and
  Pattern Recognition}, pages 9446--9454, 2018.

\bibitem{pmlr-v80-wen18a}
Haiguang Wen, Kuan Han, Junxing Shi, Yizhen Zhang, Eugenio Culurciello, and
  Zhongming Liu.
\newblock Deep predictive coding network for object recognition.
\newblock In Jennifer Dy and Andreas Krause, editors, {\em Proceedings of the
  35th International Conference on Machine Learning}, volume~80 of {\em
  Proceedings of Machine Learning Research}, pages 5266--5275,
  Stockholmsmässan, Stockholm Sweden, 10--15 Jul 2018. PMLR.

\bibitem{wen2018neural}
Haiguang Wen, Junxing Shi, Yizhen Zhang, Kun-Han Lu, Jiayue Cao, and Zhongming
  Liu.
\newblock Neural encoding and decoding with deep learning for dynamic natural
  vision.
\newblock {\em Cerebral Cortex}, 28(12):4136--4160, 2018.

\bibitem{xu2017feature}
Weilin Xu, David Evans, and Yanjun Qi.
\newblock Feature squeezing: Detecting adversarial examples in deep neural
  networks.
\newblock {\em arXiv preprint arXiv:1704.01155}, 2017.

\bibitem{feature_scatter}
Haichao Zhang and Jianyu Wang.
\newblock Defense against adversarial attacks using feature scattering-based
  adversarial training.
\newblock In {\em Advances in Neural Information Processing Systems}, 2019.

\bibitem{zhang2019theoretically}
Hongyang Zhang, Yaodong Yu, Jiantao Jiao, Eric Xing, Laurent El~Ghaoui, and
  Michael~I Jordan.
\newblock Theoretically principled trade-off between robustness and accuracy.
\newblock In {\em ICML}, 2019.

\end{thebibliography}


\newpage

\section*{SUPPLEMENTARY WORK}
This supplementary work includes: noise and detailed training descriptions as well as further examples of adversarial examples for $\varphi$DNNs.

\renewcommand{\thesection}{\Alph{section}}


\setcounter{section}{0}

\section{NOISE DESCRIPTION}
\paragraph{Panda noise.}
Given a pixel \bm{$g$} = (R, G, B) in a RGB image \textbf{I},
Panda noise can be defined as follows:
   
\begin{equation*}
    f(\bm{g}) = 
    \begin{cases}
    \bm{g},\,&P = 1-(\alpha + \beta) \\
    (255, 255, 255),\,&P = \alpha\\
    (0, 0, 0),\,&P = \beta\\
    \end{cases}
\end{equation*}
where $\alpha$ and $\beta$ are probabilities that a pixel in image \textbf{I} will become respectively white or black.

\paragraph{ColorDepth noise.}
A RGB image represents feature information by color bit depths. For example, CIFAR-10 encodes images with 24-bit color depths. The ColorDepth noise reduces original images to fewer bits representation. Given a normalized RGB image \textbf{I} that ranges from 0 to 1, and the target t-bits color depths after reducing, this noise could be formulated as:
\begin{equation*}
    \bm{I'} = \frac{[\bm{I}\cdot (2^{t} - 1)]}{2^{t} - 1}
\end{equation*}
where [] denotes the standard rounding function and \bm{$I'$} is the image encoded with t-bits color depths.
\paragraph{Gaussian noise.}
Given a RGB image \bm{$I$}, the Gaussian noise could be described as the following:
\begin{equation*}
    \bm{I'} = \bm{I} + R\sim \mathcal{N}(\mu,\,\sigma^{2})\,
\end{equation*}
where $R$ is the Gaussian filter.

\section{TRAINING STRATEGY}
We use Adam with learning scheduler in all experiments. Due to the poor performance of Adam in the adversarial training, the SGD with learning scheduler is used during the adversarial training. 
To improve neural network performance, we use learning scheduler to adjust learning rate. For Adam learning scheduler, the learning rate starts from $10^{-3}$ and increases to $10^{-1}$ then decreases to $10^{-2}$, $10^{-3}$ and $0.5*10^{-3}$. For SGD learning scheduler, the learning rate starts from $10^{-1}$ and decreases to $10^{-2}$ and $10^{-3}$.
To match data normalization, we apply sigmoid activation function instead of tanh as the last layer output for generator model in SRGAN.

\section{EXAMPLES OF IMAGES PROCESSED BY $\varphi$DNNs}

\begin{figure*}
 \centering
 \includegraphics[width=14cm,height = 20cm]{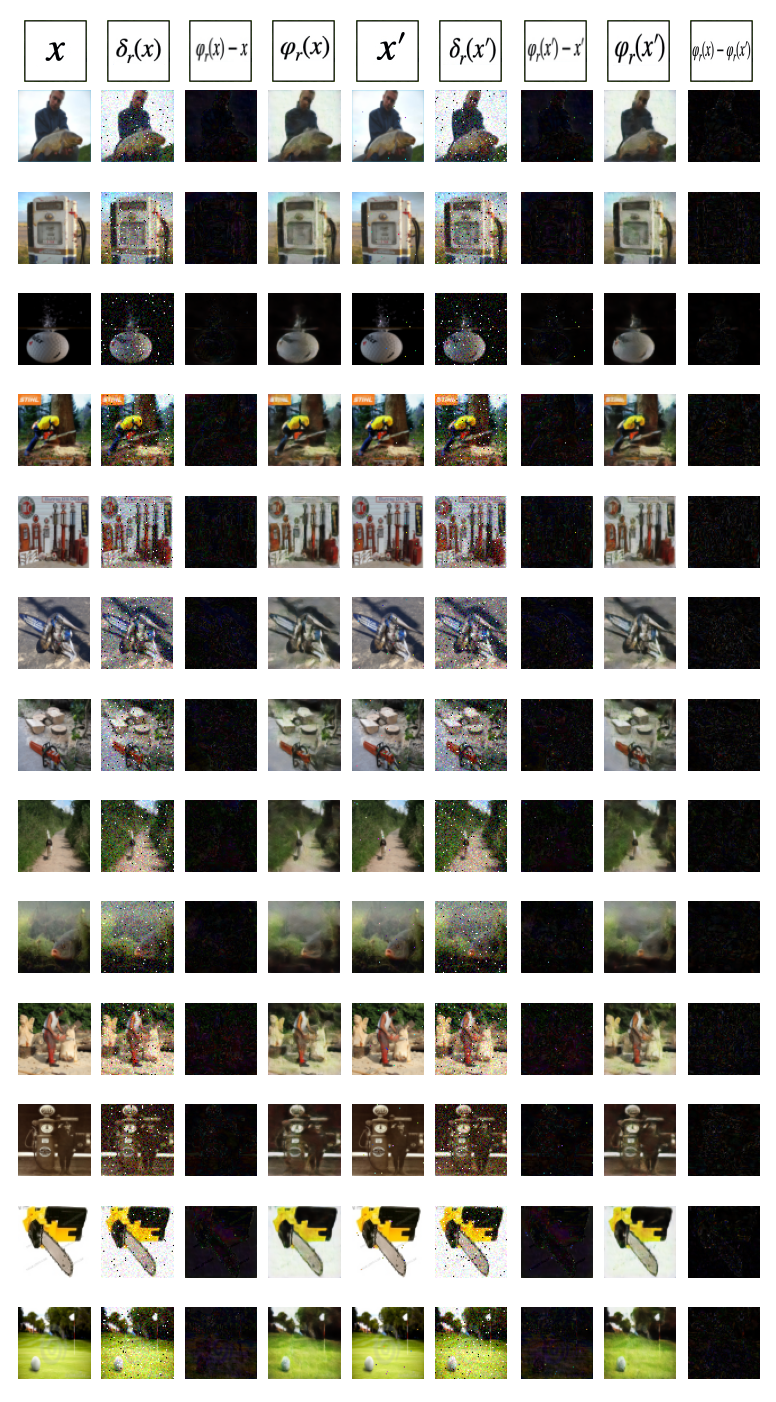}
 \caption{ Examples of $\varphi_r(x)$ for an Imagenette's image $x$ and its respective adversarial sample $x'$. The adversarial samples were built with 10px attack.}
 \label{fig:supp1}
\end{figure*}

\begin{figure*}
 \centering
 \includegraphics[width=14cm,height = 20cm]{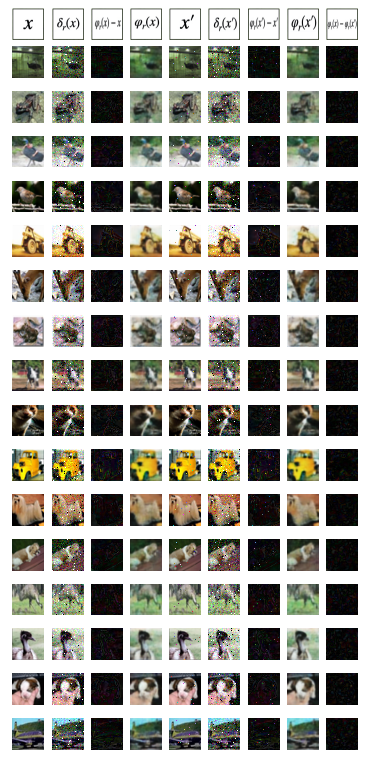}
 \caption{Examples of $\varphi_r(x)$ for an CIFAR10's image $x$ and its respective adversarial sample $x'$. The adversarial samples were built with 10px attack.}
 \label{fig:supp2}
\end{figure*}

\begin{figure*}
 \centering
 \includegraphics[width=14cm,height = 20cm]{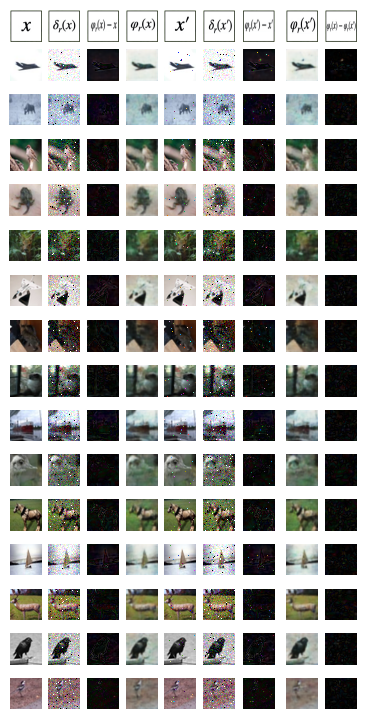}
 \caption{Examples of $\varphi_r(x)$ for an CIFAR10's image $x$ and its respective adversarial sample $x'$. The adversarial samples were built with 10px attack.}
 \label{fig:supp3}
\end{figure*}

\begin{figure*}
 \centering
 \includegraphics[width=13cm,height = 20cm]{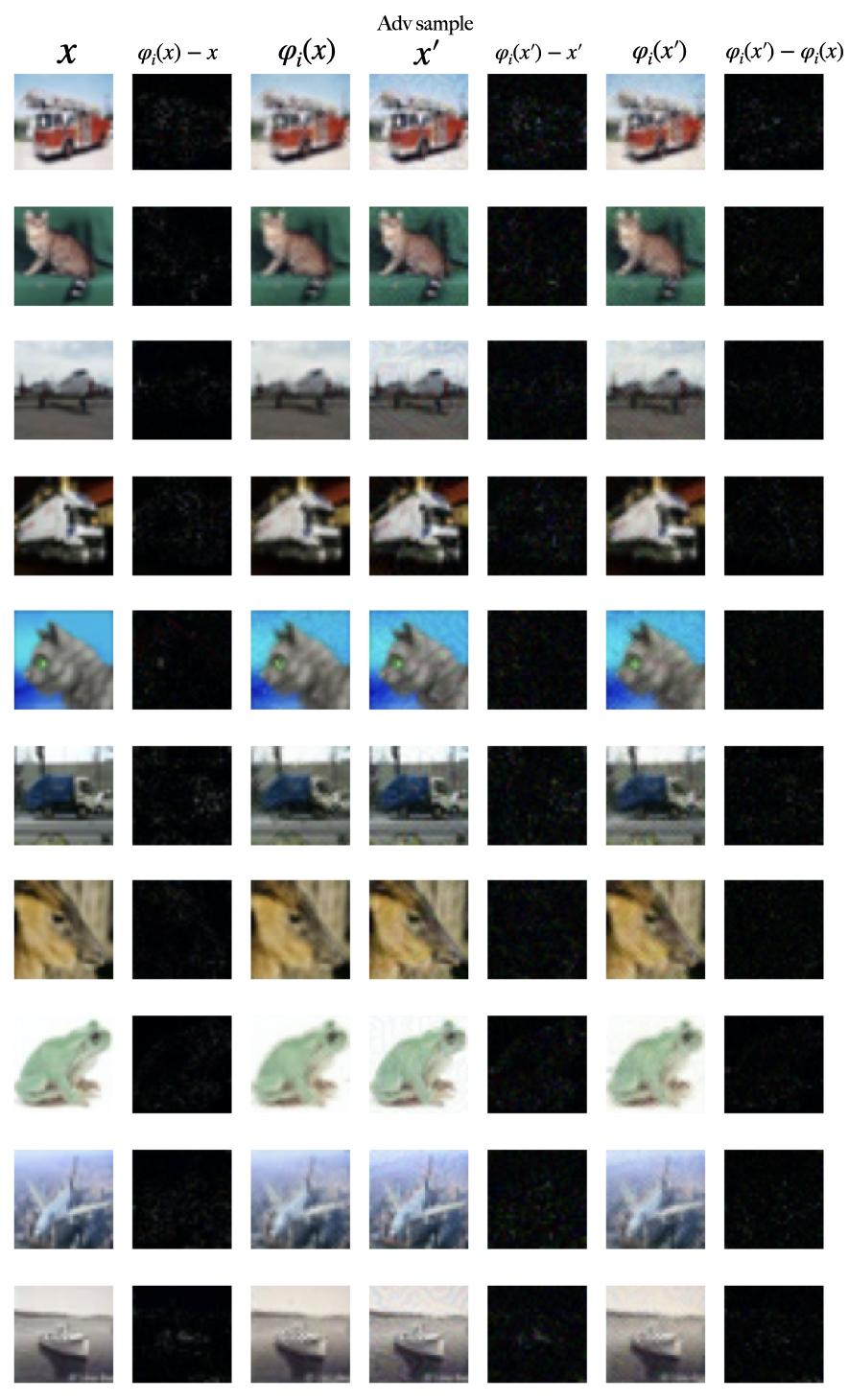}
 \caption{
 Examples of $\varphi_i(x)$ for an CIFAR10's image $x$ and its respective adversarial sample $x'$. The adversarial samples were built with PGD.
 }
 \label{fig:cifar10_pgd_supp}
\end{figure*}

\begin{figure*}
 \centering
 \includegraphics[width=13cm,height = 20cm]{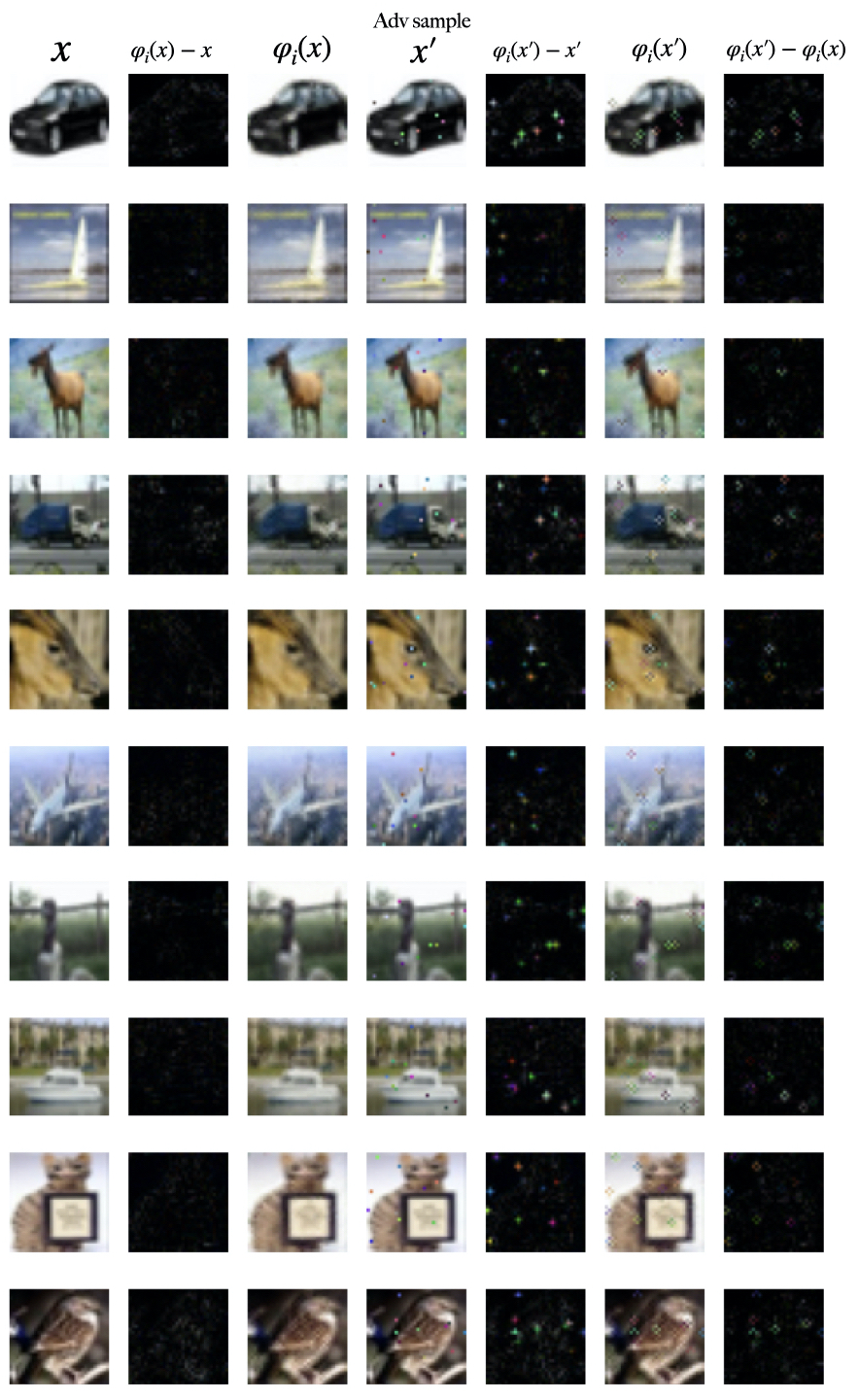}
 \caption{
  Examples of $\varphi_i(x)$ for an CIFAR10's image $x$ and its respective adversarial sample $x'$. The adversarial samples were built with 10px attack.
 }
 \label{fig:cifar10_pixel10_supp}
\end{figure*}

\begin{figure*}
 \centering
 \includegraphics[width=13cm,height = 20cm]{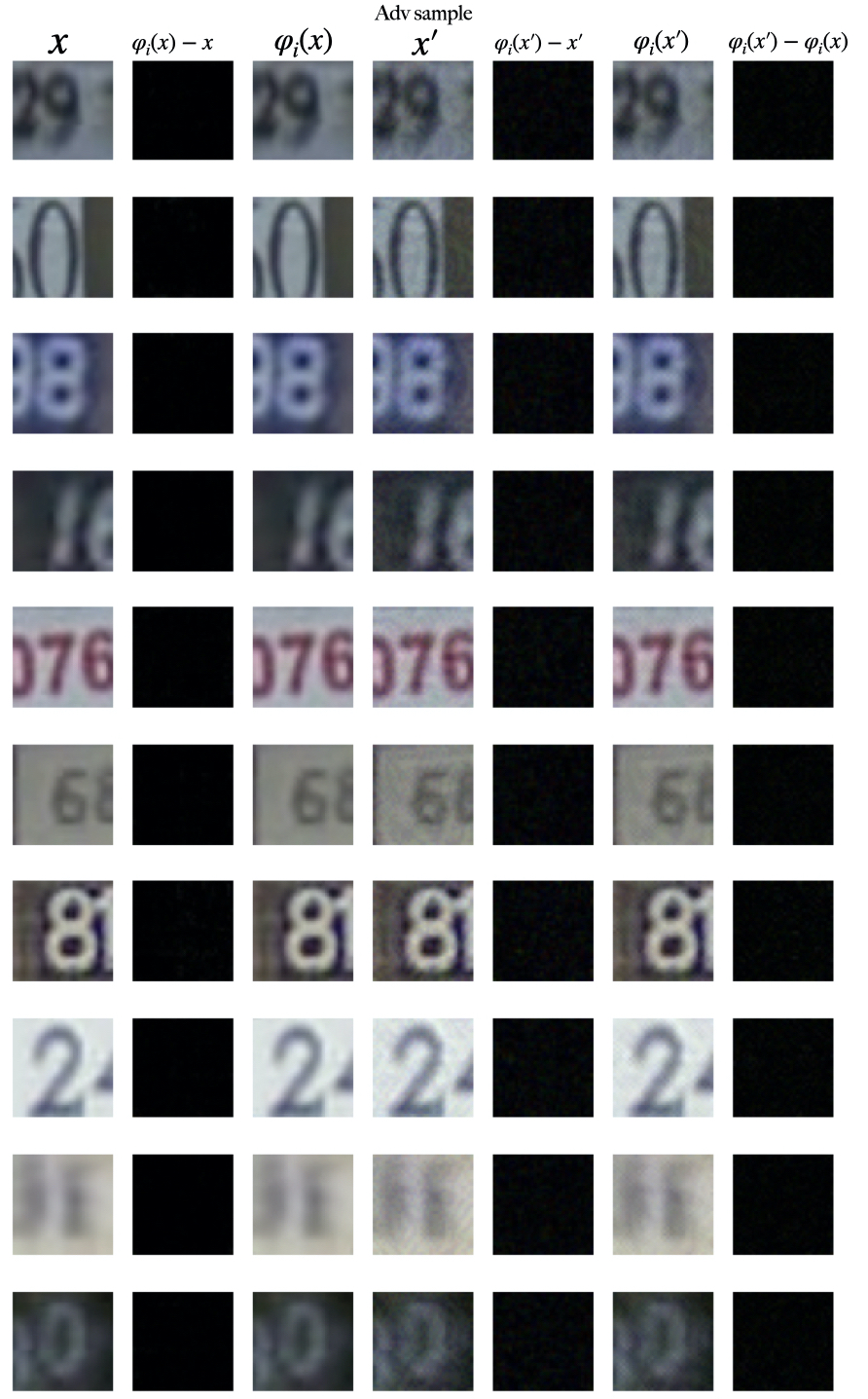}
 \caption{
 Examples of $\varphi_i(x)$ for an SVHN's image $x$ and its respective adversarial sample $x'$. The adversarial samples were built with PGD.
 }
 \label{fig:svhn_pgd_supp}
\end{figure*}

\begin{figure*}
 \centering
 \includegraphics[width=13cm,height = 20cm]{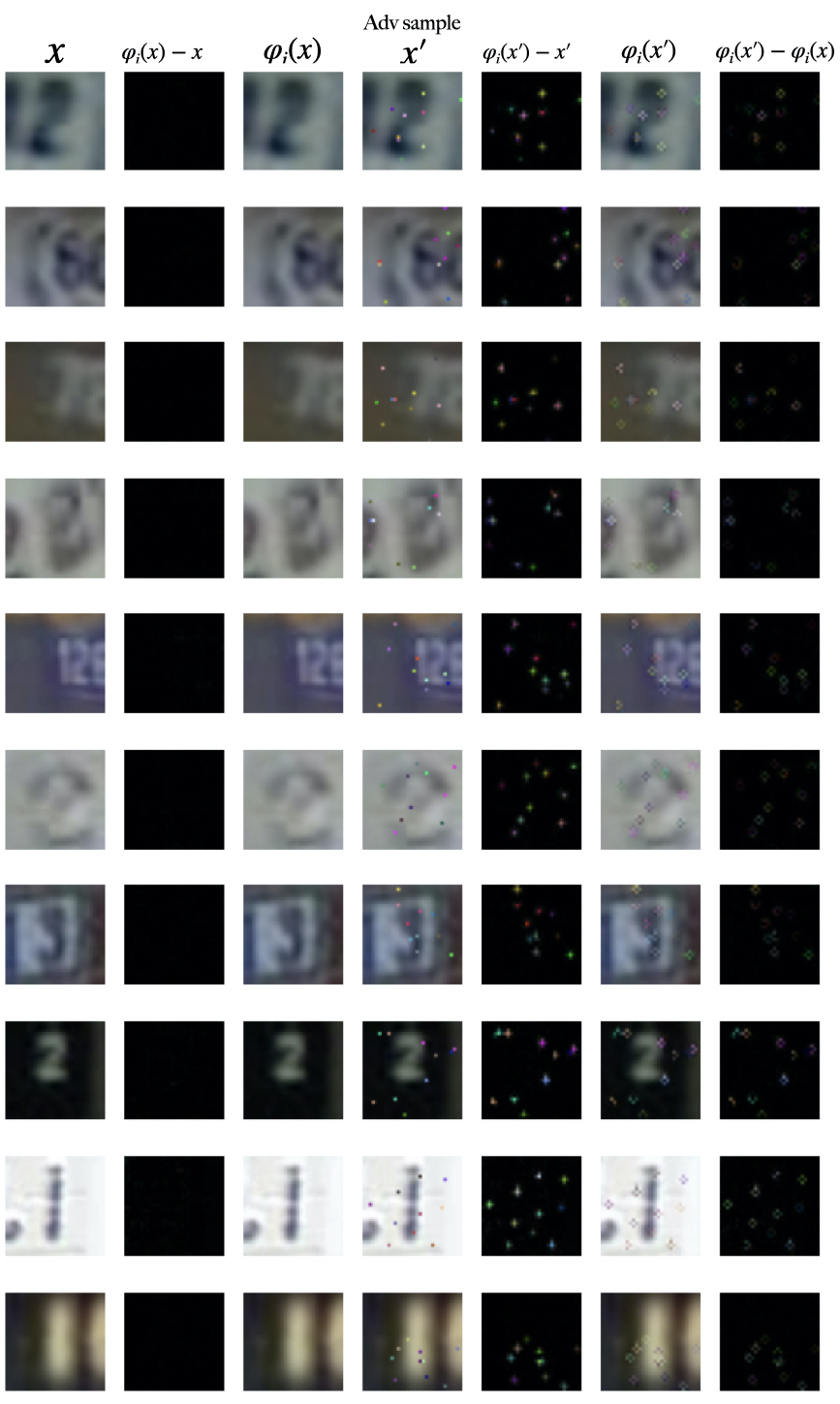}
 \caption{
  Examples of $\varphi_i(x)$ for an SVHN's image $x$ and its respective adversarial sample $x'$. The adversarial samples were built with 10px attack.
 }
 \label{fig:svhn_pixel10_supp}
\end{figure*}

\begin{figure*}
 \centering
 \includegraphics[width=13cm,height = 20cm, ]{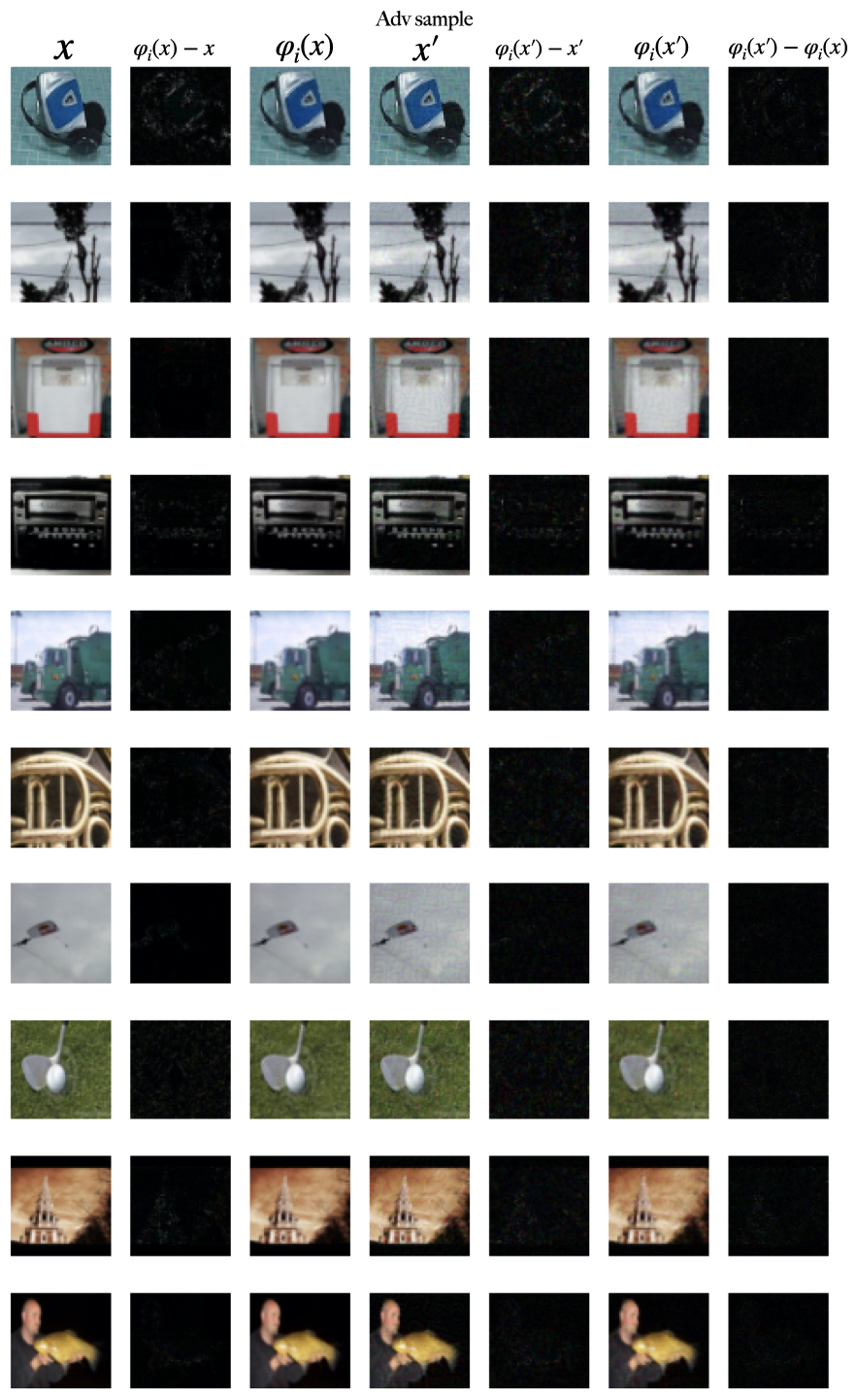}
 \caption{
 Examples of $\varphi_i(x)$ for an Imagenette's image $x$ and its respective adversarial sample $x'$. The adversarial samples were built with PGD.
 }
 \label{fig:imagenette_pgd_supp}
\end{figure*}

\begin{figure*}[b]
 \centering
 \includegraphics[width=13cm,height = 20cm]{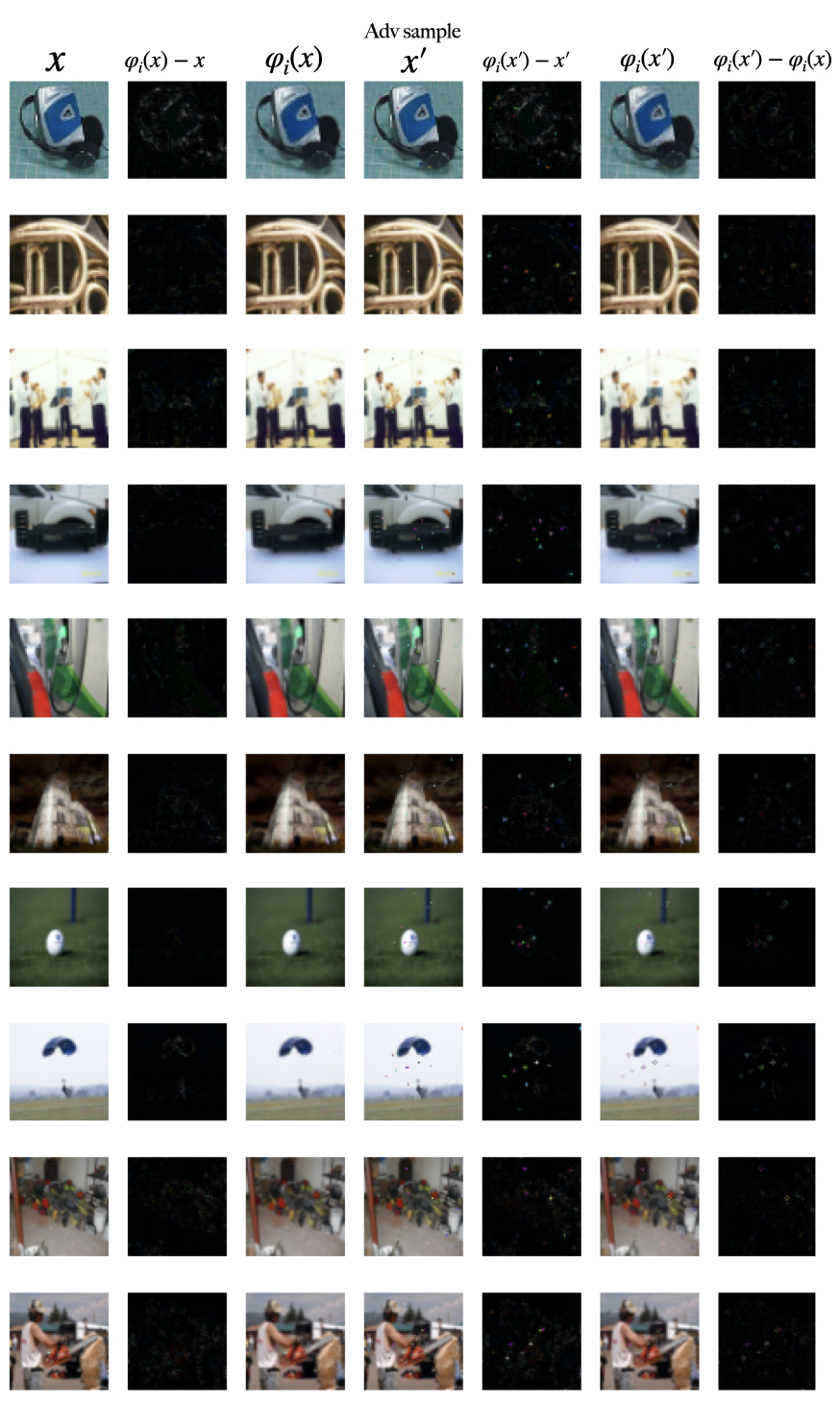}
 \caption{
 Examples of $\varphi_i(x)$ for an Imagenette's image $x$ and its respective adversarial sample $x'$. The adversarial samples were built with 10px attack.
 }
 \label{fig:imagenette_pixel10_supp}
\end{figure*}

Figure \ref{fig:supp1} to \ref{fig:supp3} contain some samples and the images from each stage of the $\varphi_r$DNN when processing these samples.
Figure \ref{fig:cifar10_pgd_supp} to \ref{fig:imagenette_pixel10_supp} follows a similar pattern but for $\varphi_i$DNN.

\end{document}